\documentclass{article}
\usepackage[preprint]{neurips_2025}
\usepackage[utf8]{inputenc}
\usepackage[T1]{fontenc}
\usepackage{hyperref}
\usepackage{url}
\usepackage{booktabs}
\usepackage{amsfonts}
\usepackage{amsmath}
\usepackage{nicefrac}
\usepackage{microtype}
\usepackage{subcaption}
\usepackage{xcolor}
\usepackage{graphicx}
\usepackage{booktabs}
\usepackage{tabularx}
\usepackage{multirow}
\usepackage{siunitx}
\usepackage{adjustbox}
\usepackage{soul} 
\newcommand{\nirmal}[1]{\textcolor{orange}{[Nirmal: #1]}}
\newcommand{\mike}[1]{\textcolor{blue}{[Mike: #1]}}
\newcommand{\amir}[1]{\textcolor{red}{[Amir: #1]}}
\newcommand{\wintext}[1]{\textbf{#1}}       

\title{Distribution-Aware Feature Selection for SAEs}

\author{
  Narmeen Oozeer\\
  Martian\\
  \texttt{narmeen@withmartian.com} \\
  \And
  Nirmalendu Prakash\\
  Singapore University of Technology and Design \\
  \texttt{email@gmail.com} \\
  \And
  Michael Lan\\
  Martian\\
  michael@withmartian.com\\
  \And
  Alice Rigg\\
  Independent\\
  \texttt{rigg.alice0@gmail.com} \\
  \And
  Amirali Abdullah\\
  Thoughtworks\\
\texttt{amir.abdullah@thoughtworks.com} \\
}

\begin{document}

\maketitle

\begin{abstract}
Sparse autoencoders (SAEs) decompose neural activations into interpretable features. A widely adopted variant, the TopK SAE, reconstructs each token from its $K$ most active latents. However this approach is inefficient, as some tokens carry more information than others. BatchTopK addresses this limitation by selecting top activations across a batch of tokens. This improves average reconstruction but risks an “activation lottery”, where rare high-magnitude features crowd out more informative but lower-magnitude ones. To address this issue, we introduce \textbf{Sampled-SAE}: we score the columns (representing features) of the batch activation matrix (via $\ell_2$ norm or entropy), forming a \emph{candidate pool} of size $K\ell$, and then apply Top-K to select tokens across the batch from the restricted pool of features. 
Varying $\ell$ traces a spectrum between batch-level and token-specific selection. At $\ell{=}1$, tokens draw only from $K$ globally influential features, while larger $\ell$ expands the pool toward standard BatchTopK and more token-specific features across the batch. Small $\ell$ thus enforces global consistency; large $\ell$ favors fine-grained reconstruction. On Pythia-160M, no single value optimizes $\ell$ all metrics: the best choice depends on the trade-off between shared structure, reconstruction fidelity, and downstream performance. Sampled-SAE thus reframes BatchTopK as a tunable, distribution-aware family.\end{abstract}

\section{Introduction}
 Sparse autoencoders (SAEs) have emerged as an essential tool for mechanistic interpretability, decomposing language model activations into sparse, interpretable features \citep{bricken2023monosemanticity, cunningham2023sparse, gao2024topk, marks2024interpreting}. The recently proposed BatchTopK SAE \citep{bussmann2024batchtopk} improves upon standard TopK by selecting features at the batch level rather than per-token, allowing variable sparsity across samples while maintaining average sparsity constraints. This modification achieves better reconstruction performance and feature activation density compared to standard TopK SAEs.

However, BatchTopK's batch-level selection still allows all features to compete equally, creating what we term an ``activation lottery''---features with rare but extreme magnitudes dominate selection over consistent mid-frequency features. Recent work \citep{sun2024hfl} shows that even high-frequency features (>10\% activation) previously dismissed as uninterpretable represent meaningful concepts such as context position. This suggests that the middle frequency range---features that fire consistently but not strongly---may be particularly valuable for interpretability but are currently underutilized due to competition from rare, high-magnitude spikes.

\begin{figure}
    \centering
    \includegraphics[width=0.95\textwidth]{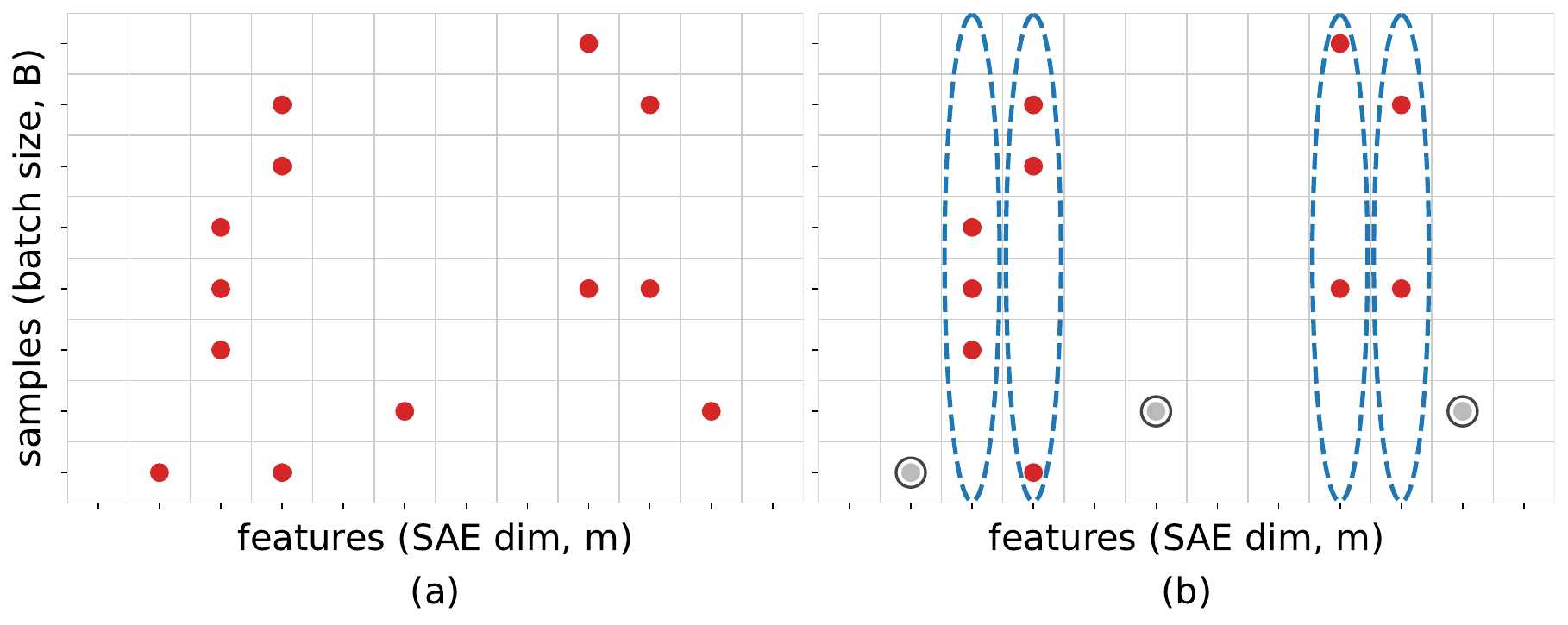}
    \caption{Comparison of BatchTopK to Sampled-SAE feature selection strategies: (a) BatchTopK selects the strongest activations across a batch, which can cause rare high-magnitude spikes to dominate, (b) Sampled-SAE introduces a candidate pool that filters out such rare spikes, ensuring more consistent feature usage across tokens.}
\label{fig:candidate_pool}
\end{figure}

We introduce a new class of Sampled-SAE that specifically promotes frequently activating features per batch. We train these on the sixth layer of Pythia-160M \citep{biderman2023pythia} using the first 25M tokens of The Pile \citep{gao2020pile}. Other training hyperparameters are in ~\ref{sae_training_config}. We also construct a controlled synthetic dataset to compare SAEs on learning disentangled features based on activation intensity and frequency. Features are partitioned based on activation frequency (low,high) and intensity (low,high), resulting in four buckets. Features are then generated by sparse superposition with small noise. Full details are in ~\ref{synthetic_data}.

We evaluate the SAEs using \textsc{SAEBench} \citep{karvonen2025saebench}—including Automated Interpretability \citep{paulo2024automatically}, $k$-sparse probing \citep{gurnee2023finding}, and Feature Absorption \citep{chanin2024absorption}. In addition, we assess: (i) coverage of high-density features (active in $\geq\!10\%$ of tokens), (ii) uniqueness of features attributable to each scoring rule, (iii) cross-seed feature similarity, and (iv) cross-scoring function similarity.

\section{Related Work}

\textbf{Sparse Autoencoders and Variants.} SAEs decompose neural network activations into interpretable features \citep{bricken2023monosemanticity, cunningham2023sparse, marks2024interpreting}. Standard TopK SAEs enforce exactly $k$ active features per token \citep{gao2024topk}, while the recent BatchTopK \citep{bussmann2024batchtopk} improves reconstruction by selecting features at the batch level, allowing variable per-token sparsity. However, BatchTopK still permits all features to compete equally, leading to what we term an ``activation lottery'' where rare, high-magnitude features dominate. Our work extends BatchTopK by introducing controlled feature pre-selection before batch-level sparsity.

\textbf{Streaming and Online Column Selection.} BatchTopK can be viewed as a streaming algorithm for feature selection, processing tokens in batches rather than requiring the full dataset. This connects to streaming matrix approximation and sketching algorithms \cite{liberty2013simple, ghashami2016frequent, woodruff2014sketching}. In streaming column subset selection (CSS), one maintains important columns while processing rows online \cite{cohen2016online, bhaskara2019online}. BatchTopK essentially performs streaming column selection where each batch determines which features (columns) to keep active. Our work extends this by adding a pre-filtering step based on batch statistics, similar to importance sampling in streaming algorithms \cite{cohen2015lp}. While BatchTopK treats all features equally in the streaming selection, we bias the selection toward features with desirable properties by restricting the candidate pool.

\textbf{Column Subset Selection and Feature Importance.} Our approach connects to the column subset selection (CSS) problem in numerical linear algebra \cite{boutsidis2009improved, mahoney2009cur}. Column L2 norms and squared L2 norms, which we employ as scoring functions, are fundamental in randomized matrix approximation \cite{frieze2004fast, drineas2006fast} and variance-based selection \cite{mahoney2009cur}. While leverage scores provide stronger theoretical guarantees \cite{mahoney2011randomized, drineas2008relative}, we focus on computationally efficient column norm methods. Future work could explore true leverage scores or ridge leverage scores \cite{cohen2017input} for SAE feature selection.

\textbf{Column Selection in Sparse Coding.} While \citet{krause2010submodular} select columns from pre-designed dictionaries (wavelets, discrete cosine transform) for sparse coding, we perform analogous selection on learned SAE features. Both approaches use greedy selection based on column importance metrics---they use variance reduction, we use column norms---to identify which dictionary elements should be available for sparse reconstruction. 

\textbf{Feature Frequency and Interpretability.} Recent work reveals that feature frequency correlates with interpretability in unexpected ways. \citet{sun2024hfl} show that high-frequency features (>10\% activation) previously dismissed as uninterpretable actually represent meaningful concepts like context position. Conversely, prior work on feature absorption \cite{chanin2024absorption} suggest that rare, high-magnitude features can dominate selection while being less interpretable. This motivates our focus on promoting consistent mid-frequency features through controlled candidate selection.

\textbf{SAE Evaluation and Alignment.} We evaluate our methods using established interpretability metrics: automated interpretability \cite{paulo2024automatically}, k-sparse probing \cite{gurnee2023finding}, and absorption \cite{chanin2024absorption}. For comparing features across different SAE configurations, we adopt the Hungarian algorithm matching approach from \citet{paulo2024sparse}, who showed that SAEs trained with different seeds learn partially overlapping feature sets.

\begin{figure}[htbp]
    \centering
    \includegraphics[width=0.95\textwidth]{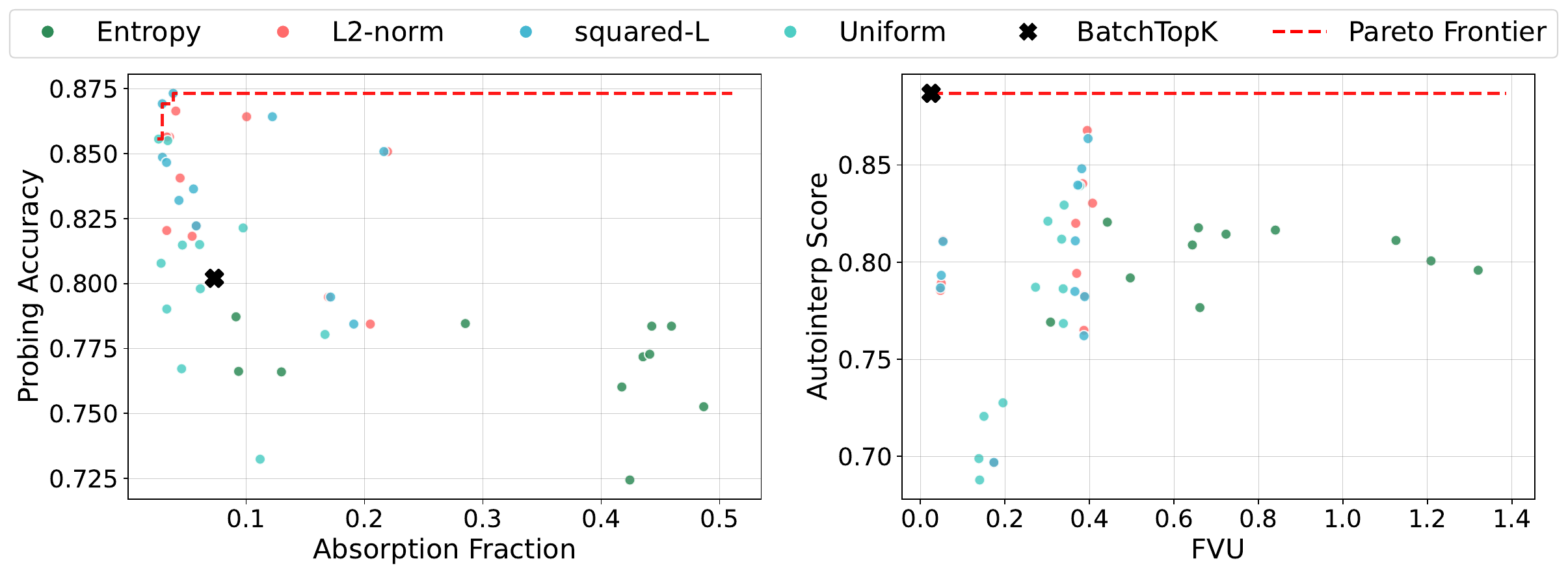}
    \caption{Pareto frontier analyses of SAE architectures: (a) Probing accuracy versus Absorption; (b) AutoInterp Scores versus FVU. While BatchTopK lies on the pareto frontier of the autointerp versus FVU curve, it is outperfrormed by the features with lower l values (Especially L2-norm and squared-L2) on the probing versus absorption fraction.}
    \label{fig:pareto-frontiers}
\end{figure}

\begin{figure}[htbp]
    \centering
    \includegraphics[width=0.95\textwidth]{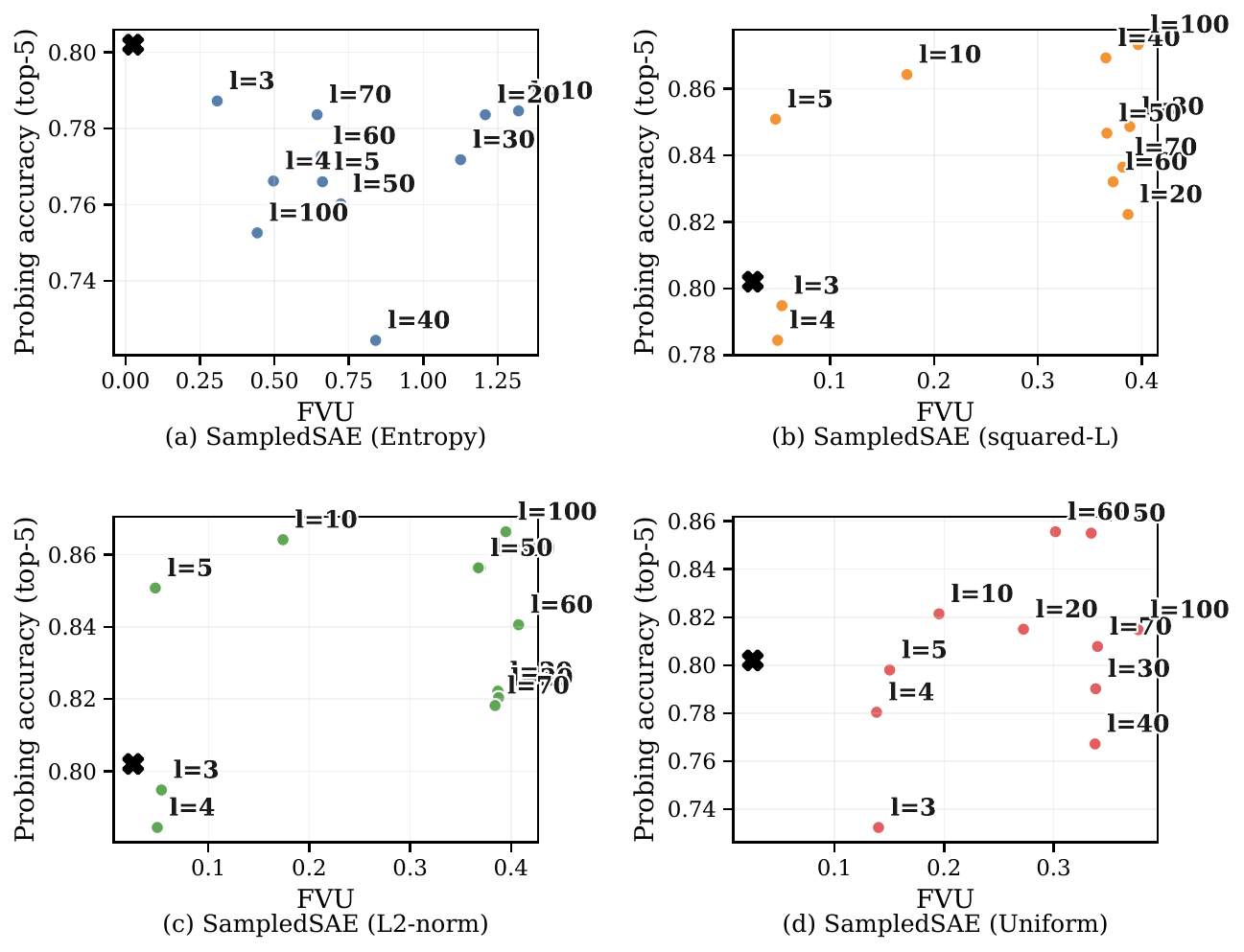}
    \caption{Probing accuracy (top-5 SAE features) vs. fraction of variance unexplained (FVU) for
SampledSAE under different candidate selection rules. Each panel shows a gating strategy with a
distinct batch-level scoring method: (a) entropy, (b) squared-L2, (c) L2-norm, and (d) uniform (random
baseline). Points correspond to different candidate set expansion factors ($\ell$).  The black cross marks the BatchTopK baseline. Higher probing accuracy and lower FVU are
both desirable: while BatchTopK performs best on reconstruction alone, distribution-aware gating
strategies (especially Squared-L2 and L2-norm) often improve probing accuracy at modest increases in
FVU.}
    \label{fig:probing_vs_fvu}
\end{figure}

\begin{figure}[htbp]
    \centering
    \includegraphics[width=0.95\textwidth]{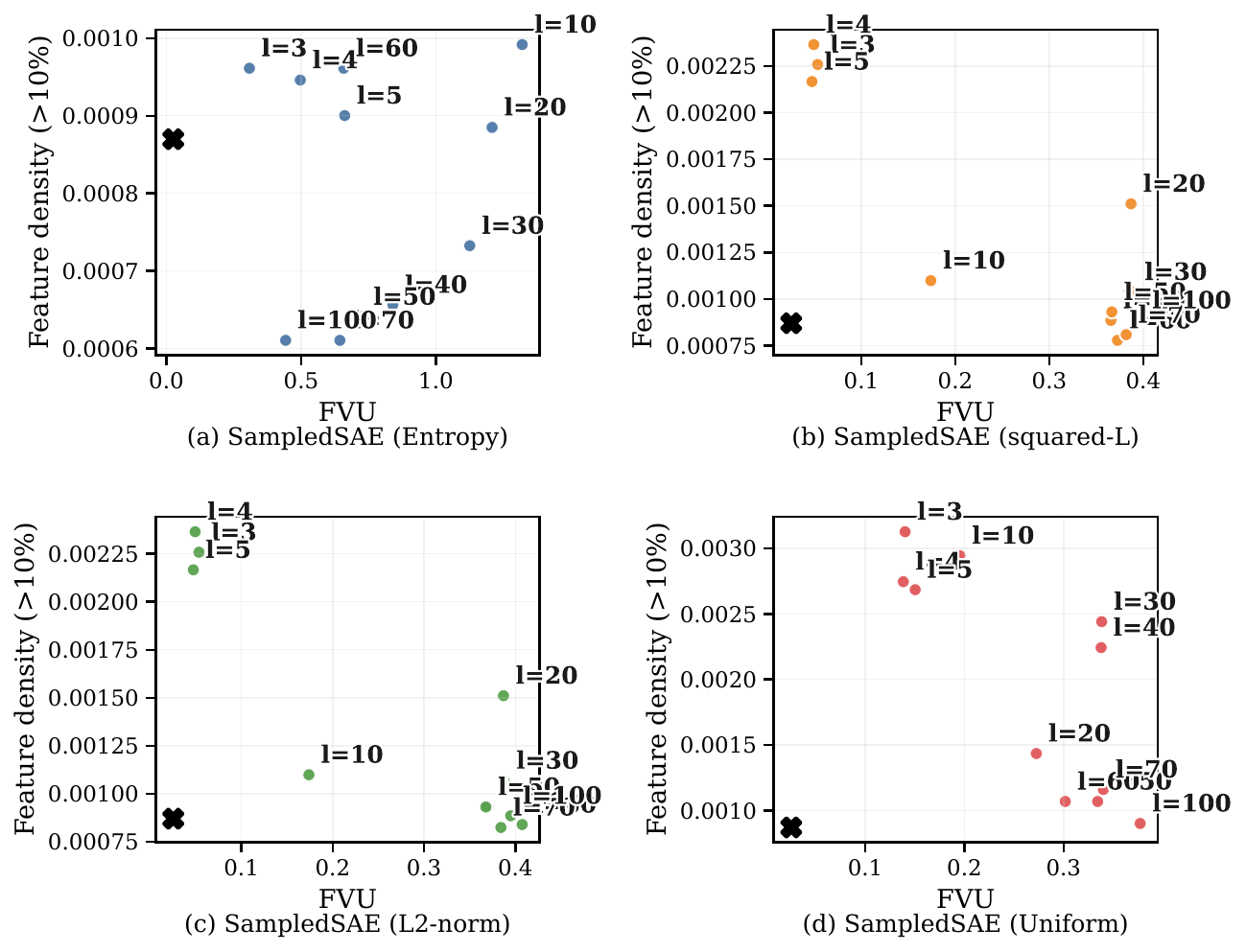}
    \caption{Fraction of SAE features active on more than 10\% of tokens vs. fraction of variance 
unexplained (FVU) under different candidate selection rules. Each panel shows a gating 
strategy with a distinct batch-level scoring method: (a) entropy, (b) Squared-L2, (c) L2-norm, and 
(d) uniform (random baseline). Points correspond to different sparsity levels ($\ell$). The black 
cross marks the BatchTopK baseline. A higher fraction indicates that more features activate 
consistently across tokens rather than only on rare spikes. Squared-L2 and L2-norm sampling strategies can achieve a much higher number of frequently activating features compared to BatchTopK.}
    \label{fig:density_vs_fvu}
\end{figure}

\begin{figure}[htbp]
    \centering
    \includegraphics[width=0.95\textwidth]{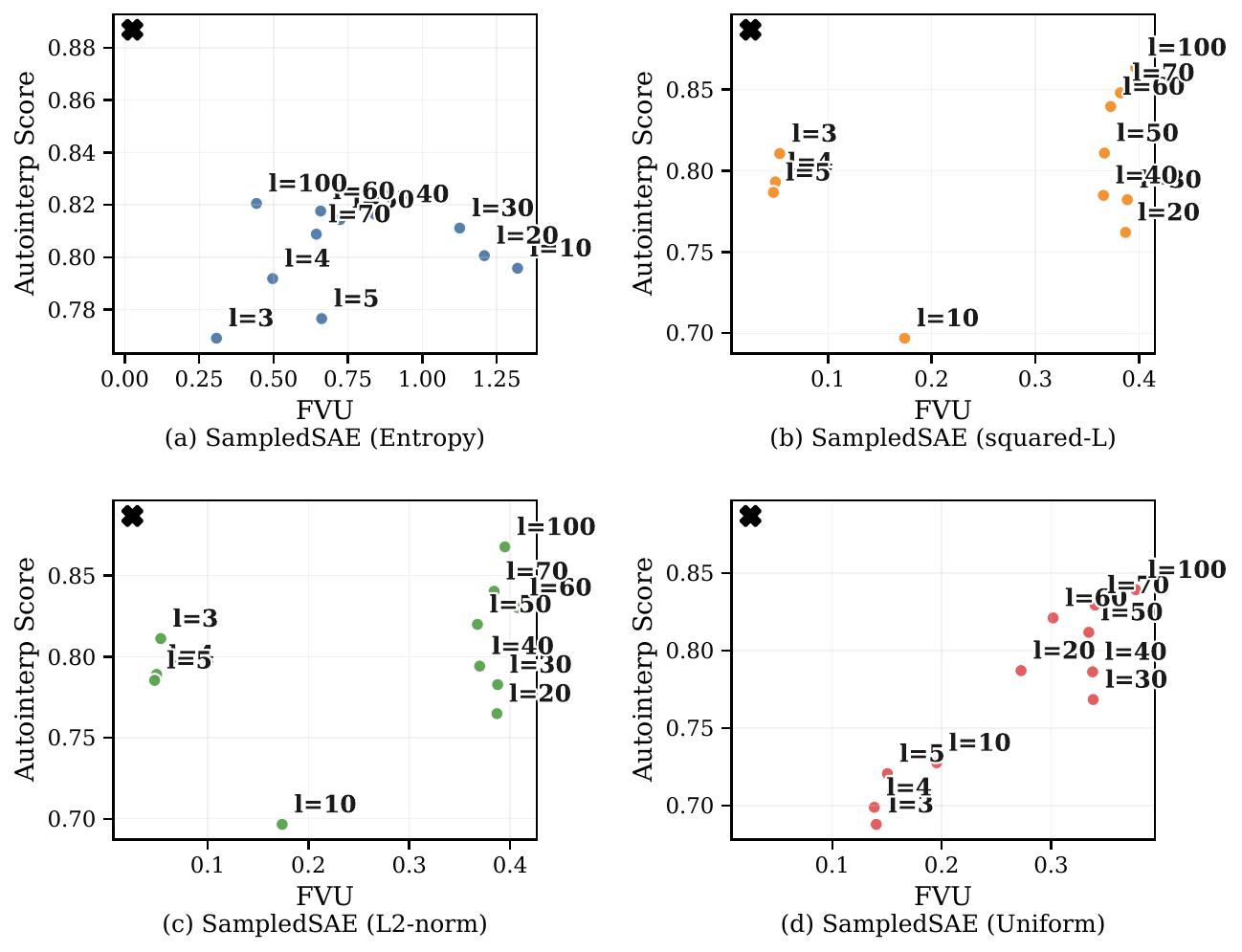}
    \caption{Automated Interpretability score vs. fraction of variance unexplained (FVU) for SampledSAE under 
    different candidate selection rules. Each panel corresponds to a different batch-level scoring rule: (a) entropy, (b) Squared-L2, (c) L2-norm, and (d) uniform 
    (random baseline). Points denote different candidate set expansion factors ($\ell$), with lower 
    values on the axis indicating better performance (lower reconstruction error) and higher value on the y-axis represent more interpretable features. The black cross marks the BatchTopK baseline. Results show that 
    distribution-aware scoring (especially Squared-L2 and L2-norm) achieves favorable trade-offs, achieving comparable auto interp scores as  BatchTopK without excessively increasing reconstruction error. We find that along these 2 axes BatchTopK is optimal.}
    \label{fig:autointerp_vs_fvu}
\end{figure}

\begin{figure}[htbp]
    \centering
    \includegraphics[width=0.95\textwidth]{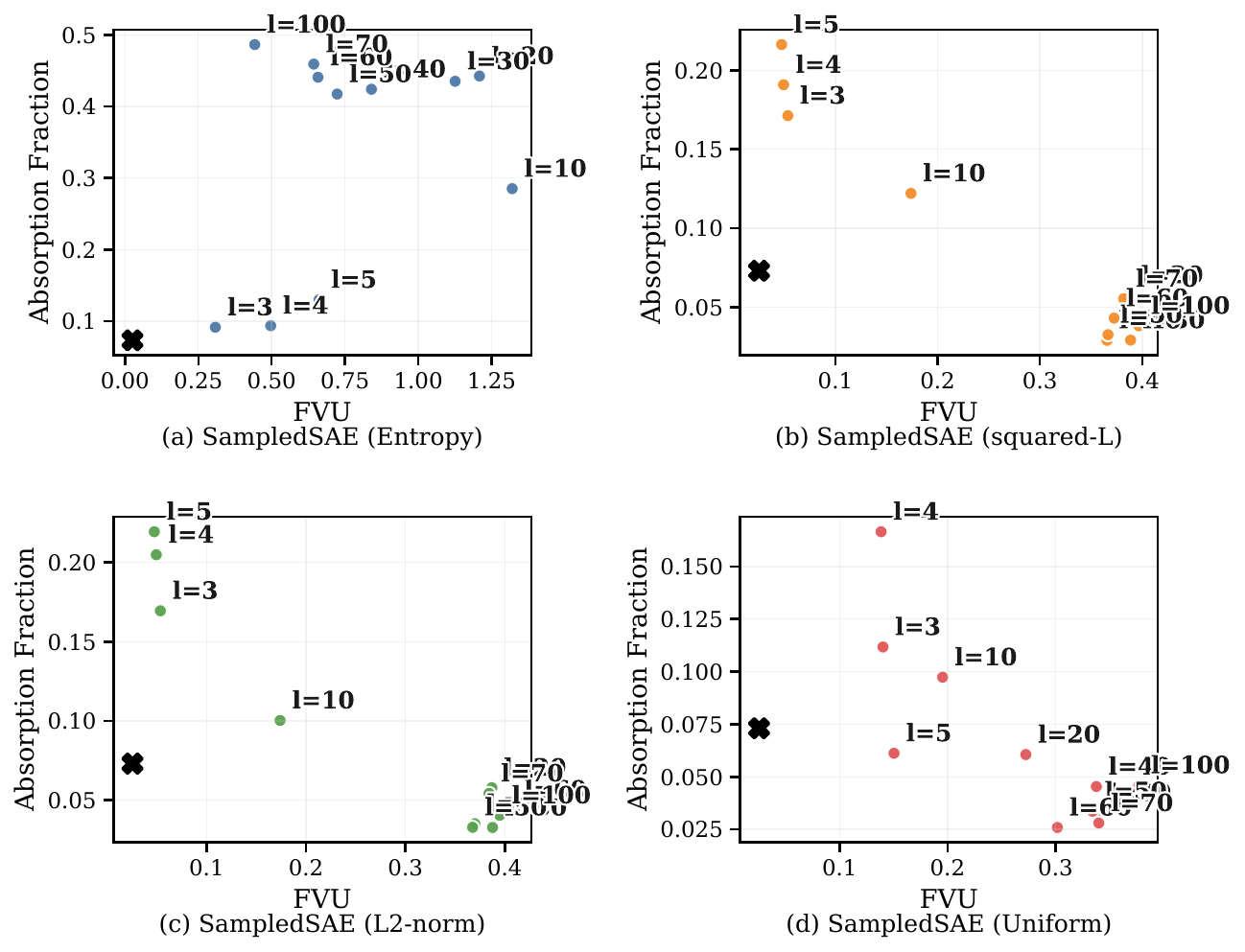}
    \caption{Absorption fraction vs. fraction of variance unexplained (FVU) for SampledSAE under 
    different candidate selection rules. Each panel corresponds to a different batch-level scoring rule: (a) entropy, (b) Squared-$\ell$, (c) L2-norm, and (d) uniform 
    (random baseline). Points denote different candidate set expansion factors ($\ell$), with lower 
    values on both axes indicating better performance (lower reconstruction error and lower absorption, 
    i.e., less feature collapse). The black cross marks the BatchTopK baseline. Results show that 
    distribution-aware scoring (especially leverage and squared-$\ell$) achieves favorable trade-offs, reducing 
    absorption without excessively increasing reconstruction error. 
    }
    \label{fig:absorption_vs_fvu}
\end{figure}

\section{Sampled Sparse Autoencoders}

We introduce Sampled-SAE, which generalizes BatchTopK through a two-stage gating process: (1) batch-level candidate selection based on feature scoring, followed by (2) Top-k sparsification within this restricted set. This decoupling enables control over which features compete for activation slots.

\textbf{Architecture.}
Sampled-SAE augments \emph{BatchTopK} with a batch-level sampling step. For a batch of activations of size $B$ and hidden dimension $n$, 
$X\!\in\!\mathbb{R}^{n\times B}$ compute encoder preactivations Z.

\begin{equation}
Z = W_{\mathrm{enc}} X + b_{\mathrm{enc}} \in \mathbb{R}^{m\times B}.
\label{eq:preacts}
\end{equation}

The m columns of Z represent features and the B rows represent input samples within the batch. Score features over the batch with a global scoring function $s_{\phi}$ (e.g., squared-L2, entropy, or $\ell_2$), produce
\begin{equation}
q = s_{\phi}(Z) \in \mathbb{R}^{m},
\label{eq:score}
\end{equation}
and select a candidate pool using feature scoring functions defined in ~\ref{sec:scoring_functions} of size $K\ell$ via
\begin{equation}
c = \operatorname{TopK}\!\big(q,\, K\ell\big) \in \{0,1\}^{m}.
\label{eq:pool}
\end{equation}

Mask the preactivations with the broadcasted pool $c\,\mathbf{1}_B^\top$ and apply \emph{BatchTopK} to obtain sparse codes
\begin{equation}
F = \operatorname{BatchTopK}\!\big(Z \odot (c\,\mathbf{1}_B^\top),\, K\big)\in\mathbb{R}^{m\times B}.
\label{eq:codes}
\end{equation}

then reconstruct and train with
\begin{equation}
\hat X = F W_{\mathrm{dec}} + b_{\mathrm{dec}},
\label{eq:recon}
\end{equation}
\begin{equation}
\mathcal{L}_{\text{Sampled}}(X) = \lVert X - \hat X\rVert_F^2 + \alpha\,\mathcal{L}_{\text{aux}}(F).
\label{eq:loss}
\end{equation}


\textbf{Candidate Selection}: We first compute batch-level scores $s_j$ for each feature $j$ and select the top $m = \lfloor \ell \cdot k \rfloor$ features to form candidate set $S$. The candidate pool expansion factor $\ell\in[1,\,n/K]$ controls the degree of filtering:
\begin{itemize}
\item $\ell = 1$: Only $k$ features compete (most restrictive)
\item $\ell = n/k$: All features compete (recovers BatchTopK)
\item $1 < \ell < n/k$: Partial filtering of rare features
\end{itemize}

Crucially, smaller $\ell$ values prevent rare, high-magnitude features from entering the candidate pool, promoting consistent mid-frequency features instead. BatchTopK is thus a special case where $\ell = n/k$, allowing all features including rare spikes to compete.

\if false
\textbf{Row-wise Sparsification}: Next, given restricted activations $Z' = Z \odot M^{\top}$ where $M_j = \mathbf{1}[j \in S]$, we apply standard Top-k per token:
\begin{equation}
J_i = \text{arg-top-k}(Z'_{i,:}, k), \quad (Z_{\text{sparse}})_{i,j} = \begin{cases} Z'_{i,j} & j \in J_i \\ 0 & \text{otherwise} \end{cases}
\end{equation}
\fi 

\subsection{Feature Scoring Functions}
\label{sec:scoring_functions}
We evaluate four scoring strategies that capture different notions of feature importance:

\textbf{$\ell_2$-norm:} $s_j = \|Z_{:,j}\|_2$ computes the column $\ell_2$ norm as a fast approximation to leverage scores from randomized matrix theory. This emphasizes features with consistent batch-level activation---a feature firing at magnitude 10 across 50\% of samples scores higher than one firing at magnitude 100 on 5\% of samples. This rewards stability over sporadic extreme activations.

\textbf{Entropy:} $s_j = -\sum_{b=1}^B q_{bj} \log(q_{bj} + \epsilon)$ where $q_{bj} = Z_{bj}/\sum_{b'} Z_{b'j}$ prefers features with selective firing patterns. Low entropy indicates a feature concentrates its activation on specific inputs rather than firing uniformly. This rewards specialization---features that strongly activate for particular contexts while remaining quiet elsewhere.

\textbf{Squared-L2:} $s_j = \sum_{b=1}^B Z_{bj}^2 + \lambda$ rewards total activation energy across the batch. Unlike the L2-norm, Squared-L2 uses squared magnitudes, making it sensitive to both frequency and high intensity. This favors features that either fire frequently at moderate strength or occasionally at very high strength. The ridge term $\lambda$ provides numerical stability.

\textbf{Uniform:} $s_j = \text{const}$ assigns equal scores to all features, resulting in random sampling. This baseline tests whether structured selection based on batch statistics improves over random feature subsampling. Even random sampling with small $\ell$ can improve some metrics by excluding rare features, though it lacks the systematic benefits of informed selection.

Each strategy creates different biases: L2-norm and Squared-L2 favor consistent mid-frequency patterns, entropy selects specialized features, and uniform provides an unbiased baseline. These differences become more pronounced at smaller $\ell$ values.




\section{Synthetic Data Experiments}

To understand how different sampling strategies affect feature recovery across the frequency-magnitude spectrum, we design controlled experiments with ground-truth features. We generate data $X \in \mathbb{R}^{n \times d}$ through sparse linear combinations of $k > d$ dictionary features: $X = SA^T + \epsilon$, where $A \in \mathbb{R}^{d \times k}$ has minimized mutual coherence, $S \in \mathbb{R}^{n \times k}$ contains sparse codes, and $\epsilon$ is 20dB SNR noise. We use $n=10,000$ samples, $d=256$, $k=1024$, forcing superposition since $k > d$.

We partition features equally into four categories based on activation frequency ($p$) and magnitude ($\sigma$): LF+HA ($p=0.02$, $\sigma=1.0$), HF+HA ($p=0.20$, $\sigma=1.0$), LF+LA ($p=0.02$, $\sigma=0.2$), and HF+LA ($p=0.20$, $\sigma=0.2$). Each feature's support follows Bernoulli($p$) with coefficients drawn from $\mathcal{N}(0, \sigma^2)$. This creates an ``activation lottery'' scenario where rare high-magnitude features (LF+HA) would typically dominate selection in standard BatchTopK. The dataset statistics are covered in \ref{synthetic_data}.

We train both Sampled-SAEs and a BatchTopK-SAE on the generated dataset. The BatchTopK-SAE learns effectively, achieving a fraction of variance explained (FVE) of \(\approx 0.99\), while Sampled-SAEs perform substantially worse (see training plots in ~\ref{synthetic_sae_training}). We are actively troubleshooting Sampled-SAE training instabilities. Once satisfactory reconstructions are obtained, we will evaluate:
\begin{enumerate}
    \item Bucket recovery rate — fraction of ground-truth features successfully recovered (match similarity $\geq 0.7$), using Hungarian matching following \citep{paulo2025sparse}.
    \item Activation fidelity — correlation between learned and true activation frequencies.
\end{enumerate}




\section{Real Data Experiments}

We train a class of SAEs parametrized by $\ell$ where BatchTopK is a special case with $\ell = n/k$, allowing us to evaluate the full spectrum from aggressive filtering ($\ell = 1$) to unrestricted selection. Our implementation extends the dictionary learning codebase from \citet{marks2024dictionary_learning}\footnote{\url{https://github.com/saprmarks/dictionary_learning}}, adding the candidate selection mechanism described in Section 3.1. We train on the sixth layer of Pythia-160M [Biderman et al., 2023] using the first 25M tokens of The Pile [Gao et al., 2020], with full training configurations detailed in Table~\ref{tab:training-config}. We evaluate each configuration along multiple interpretability axes using SAEBench \citet{karvonen2025saebench} , as described below.

\textbf{Feature Density ($>10\%$):} Following \citet{sun2024hfl} showing that high-frequency features ($>10\%$ activation) represent meaningful concepts like context position, we measure the proportion of SAE features that activate on more than $10\%$ of tokens. SampledSAE variants, particularly L2-norm and Squared-$\ell$ scoring strategies, achieve substantially higher feature densities compared to BatchTopK (Figure \ref{fig:density_vs_fvu}). This suggests that distribution-aware candidate selection promotes consistent mid-frequency features over rare high-magnitude spikes. L2-norm and Squared-L2 can achieve 2-3$\times$ higher densities of frequently activating features while maintaining comparable reconstruction fidelity, indicating that the ``activation lottery'' problem in BatchTopK systematically underutilizes dense features.

\textbf{Sparse Probing:} Sparse probing evaluates whether SAEs isolate pre-specified concepts by identifying the $k$ most relevant latents for each concept (e.g., sentiment) through comparing their mean activations on positive versus negative examples, then training linear probes on these top-$k$ latents. We find that SampledSAEs with lower $\ell$ values ($\ell = 3$-$5$) achieve better probing accuracy than BatchTopK while maintaining comparable reconstruction fidelity (Figure \ref{fig:probing_vs_fvu}). L2-norm and Squared-$\ell$ sampling strategies consistently outperform BatchTopK on this metric, likely because they promote consistent mid-frequency features over rare high-magnitude spikes that may represent less generalizable concepts. Entropy sampling underperforms BatchTopK, while uniform sampling with high $\ell$ values trades reconstruction quality for improved concept detection.

\textbf{Feature Absorption:} Feature absorption is a phenomenon where sparsity incentivizes SAEs to learn undesirable feature representations with hierarchical concepts where A implies B---rather than learning separate latents for both concepts, the SAE learns a latent for A and a latent for ``B except A'' to improve sparsity. Our results show that distribution-aware sampling strategies, particularly L2-norm and Squared-$\ell$, substantially reduce absorption compared to BatchTopK (Figure 6). This improvement suggests that preventing rare high-magnitude features from dominating selection helps maintain cleaner concept boundaries and reduces the gerrymandered latent patterns characteristic of feature absorption.

\textbf{Automated Interpretability:} Automated interpretability \citet{paulo2024automatically} uses an LLM-based judging framework where a language model first proposes a ``feature description'' using activating examples, then constructs test sets by sampling sequences across different activation strengths along with control sequences, with the LLM judge using the feature description to predict which sequences would activate the latent. While BatchTopK performs well on this metric, SampledSAE variants achieve comparable interpretability scores with only modest increases in reconstruction error (Figure \ref{fig:autointerp_vs_fvu}). The similar performance across scoring strategies suggests that the automated interpretability metric may be less sensitive to the specific feature selection mechanism than task-specific metrics like probing and absorption.

\textbf{Main Takeaway:} The Pareto frontier analysis across different metric pairs reveals that no single SAE configuration dominates across all interpretability dimensions (Figure \ref{fig:pareto-frontiers}). While BatchTopK lies on the Pareto frontier for automated interpretability versus reconstruction fidelity (FVU), it is dominated by SampledSAE variants on the probing accuracy versus absorption frontier. L2-norm and Squared-$\ell$ scoring strategies achieve superior positions on the probing-absorption trade-off, simultaneously improving concept detection while reducing feature absorption—a combination that BatchTopK's unrestricted selection cannot achieve. This demonstrates that SAE selection requires careful consideration of the specific interpretability goals, as optimizing for reconstruction fidelity alone misses important advantages in concept isolation and feature disentanglement.

We also compute additional metrics to compare our different scoring strategies:

\textbf{Unique Features Discovered by SampledSAEs:} To identify features uniquely captured by each architecture, we compute cross-architecture similarity between SampledSAE variants and BatchTopK features. For each feature pair $(i, j)$, we measure similarity through two channels: decoder similarity $s_{\text{dec}}(i, j)$, computed as the cosine similarity between L2-normalized decoder weight vectors, and semantic similarity $s_{\text{text}}(i, j)$, based on the automated natural language explanations generated for each feature by Automated Interpretability \citet{paulo2024automatically}. These explanations describe what concept each feature represents, which we embed using Sentence-Transformers (all-mpnet-base-v2) and compare via cosine similarity. These metrics are combined into an overall similarity score $s_{\text{comb}}(i, j) = (s_{\text{dec}}(i, j) + s_{\text{text}}(i, j))/2$, where $s_{\text{comb}} \in [-1, 1]$. For each feature $i$ in one architecture, we identify its best match $j^* = \arg\max_j s_{\text{comb}}(i, j)$ in BatchTopK SAE. Features with low best-match scores represent concepts uniquely captured by that SAE variant—they lack strong equivalents in the comparison architecture. 

Tables~\ref{tab:features unique to Entropy}, \ref{tab:features unique to L2}, \ref{tab:features unique to Squared-L}, and \ref{tab:features unique to Uniform} show the most unique features discovered by each SampledSAE variant (Entropy, L2-norm, Squared-$\ell$, and Uniform) compared to BatchTopK, with best-match similarities typically below 0.15. These low similarity scores demonstrate that controlled candidate selection not only changes which features are selected but discovers genuinely different interpretable structures in the activations. The selection strategies exhibit distinct biases in their highest-quality features (Tables~\ref{tab:top10-batchtopk}--\ref{tab:top10-uniform}): BatchTopK learns a mix of abstract domain-specific concepts and high-frequency tokens, L2-norm and Squared-$\ell$ favor frequent compositional structures (variables, HTML tags, relational operators), while Entropy captures discriminative patterns across granularities and Uniform defaults to syntactic elements. These differences suggest each scoring function implicitly selects for different computational roles features play in the network (see ~\ref{app:feature-analysis} for detailed analysis).

\textbf{Feature Similarity Across Seeds}:
SAEs are known to exhibit seed variance, learning different features under different random initializations \citep{paulo2025sparse}. We train Sampled-SAE with three seeds for each sampling strategy and quantify cross-seed agreement using mean max cosine similarity (MMCS; the mean, over features, of each feature's maximum cosine match \citep{braun2025interpretability}) across seeds. BatchTopK shows markedly higher cross-seed agreement (than the other strategies (Table~\ref{tab:mmcs_sampling}), indicating substantially more consistent features across seeds.

\textbf{Feature Similarity Across Scoring Functions}: Using BatchTopK as the reference, we obtain a MMCS of $\approx 0.72$ when matching learned features against each scoring-based SAEs.

\begin{table}[t]
\centering
\caption{Mean max cosine similarity (MMCS) for SAEs trained with different sampling methods and seeds(0,1, and 2). Base = seed 0; compared against seeds 1 and 2. Reported as mean over the two comparisons ($n=2$); standard deviation is close to zero.}
\label{tab:mmcs_sampling}
\begin{tabular}{lrr}
\toprule
\textbf{Sampling method} & MMCS\\
\midrule
Entropy   & 0.176 \\
Squared-$\ell$   & 0.176 \\
$\ell_2$-norm  & 0.179 \\
Uniform   & 0.157 \\
BatchTopK & \textbf{0.277} \\
\bottomrule
\end{tabular}
\end{table}

\section{Discussion and Conclusion}
SampledSAE provides a general framework for studying how features are selected in sparse autoencoders. By decoupling candidate selection from row-wise Top-$k$, we show that distribution-aware gating improves utilization and interpretability. Limitations include the added hyperparameter $s$ and reliance on batch-level statistics. Future directions include adaptive scoring functions, learned gating mechanisms, and connections to mixture-of-experts routing.

\section*{Limitations}

\begin{itemize}
    \item \textbf{Unproven interpretability of mid-activation, high-frequency features.}  
    While we hypothesize such features might be interpretable compared to spiky features, we do not contrast interpretability of these features vis-a-vis high activating and less dense features to fully validate this hypothesis.

    \item \textbf{Lack of theoretical guarantees.}  
    We do not provide mathematical proofs that SampledSAE yields better feature approximations than BatchTopK, nor adversarial cases where BatchTopK performs poorly.

    \item \textbf{Trade-off between density and interpretability.}  
    SampledSAE recovers denser features that improve probing accuracy and reduce feature absorption scores, but we find no evidence that this improves interpretability.

    \item \textbf{Reliance on autointerp as a proxy.}  
    We evaluate interpretability primarily through autointerp, while noting its known limitations.

    \item \textbf{Unproven choice of scoring function.} 
    We find no strong justification for any particular scoring function. Our experiments bias toward norm- and entropy-based measures, while we hypothesize that functions with favorable geometric properties could lead to more unique features and low-dimensional approximations of feature space geometry. Developing theory for such sampling strategies remains future work.

    \item \textbf{Restricted scope of training.}  
    We train SAEs only on a single layer of a single model model due to computational constraints, limiting the generality of our findings.
    
    \item \textbf{Introduction of a new hyperparameter.} 
    The axis $\ell$ enables exploration of trade-offs relevant to interpretability, but adds another hyperparameter to optimize.

\end{itemize}
\bibliography{neurips}
\appendix


\section{Appendix}
\subsection{Synthetic Data}
\label{synthetic_data}
\begin{table}[h!]
\centering
\caption{Synthetic dataset summary. Overall: mutual coherence $=0.0833$ (lower is better); expected $L_0$ per sample $=61.52$; observed $L_0$ per sample $=61.51$. Bucket-level mean active coefficient magnitude is reported with standard deviation in parentheses.}
\label{tab:synthetic_stats}
\begin{tabular}{lrr}
\toprule
\textbf{Bucket} & \textbf{\#Features} & \textbf{Mean activation $\pm$ std} \\
\midrule
LF+HA & 265 & $0.801 \pm 0.605$ \\
HF+HA & 237 & $0.798 \pm 0.601$ \\
LF+LA & 288 & $0.159 \pm 0.120$ \\
HF+LA & 234 & $0.159 \pm 0.120$ \\
\bottomrule
\end{tabular}
\end{table}

\subsection{Synthetic SAE Training}
\label{synthetic_sae_training}

\begin{figure*}[h!]
  \centering
  \begin{subfigure}[t]{0.5\textwidth}
    \centering
    \includegraphics[width=\linewidth]{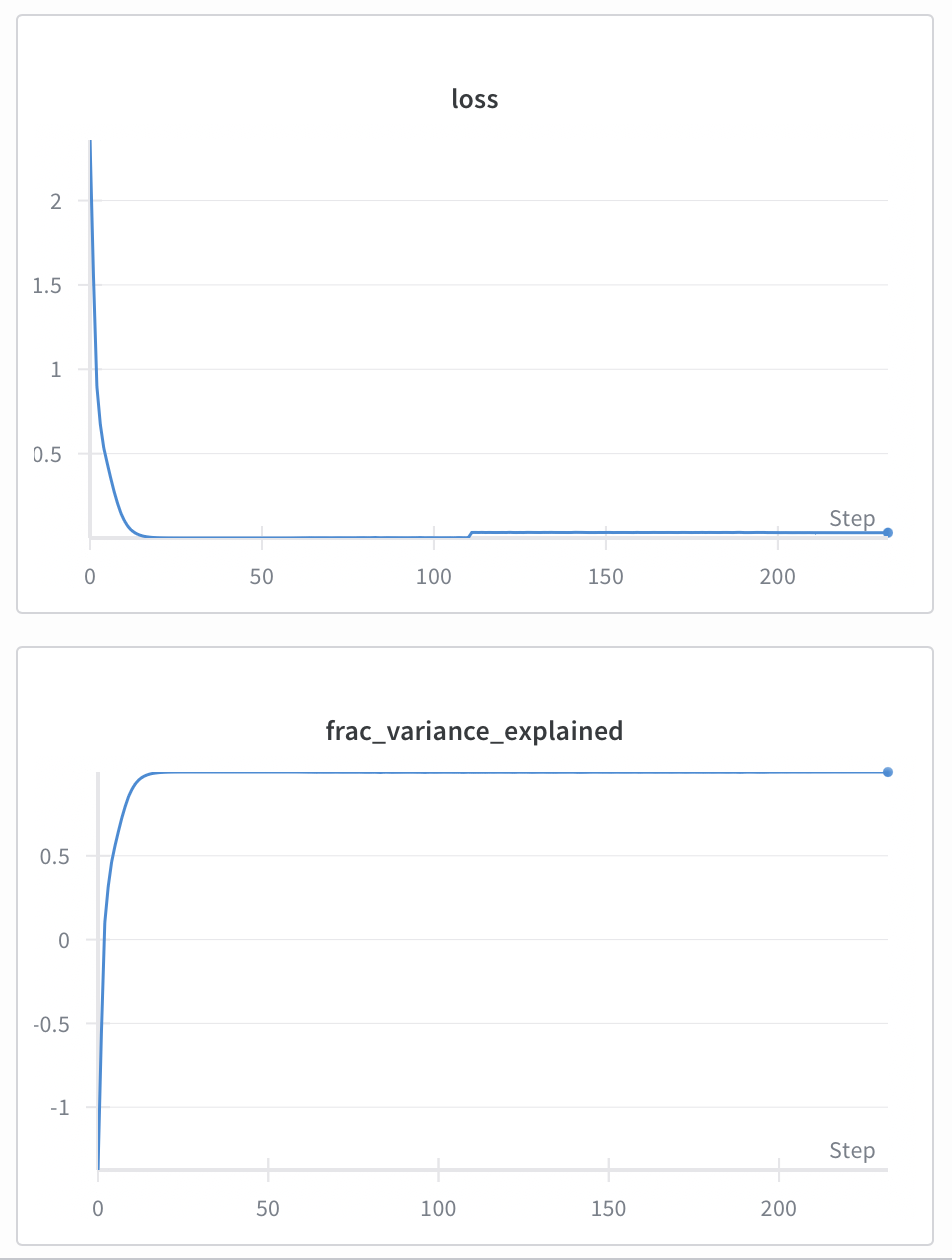}
    \caption{Loss and FVE for BatchTopK on synthetic data}
    \label{fig:left}
  \end{subfigure}
  \begin{subfigure}[t]{0.5\textwidth}
    \centering
    \includegraphics[width=\linewidth]{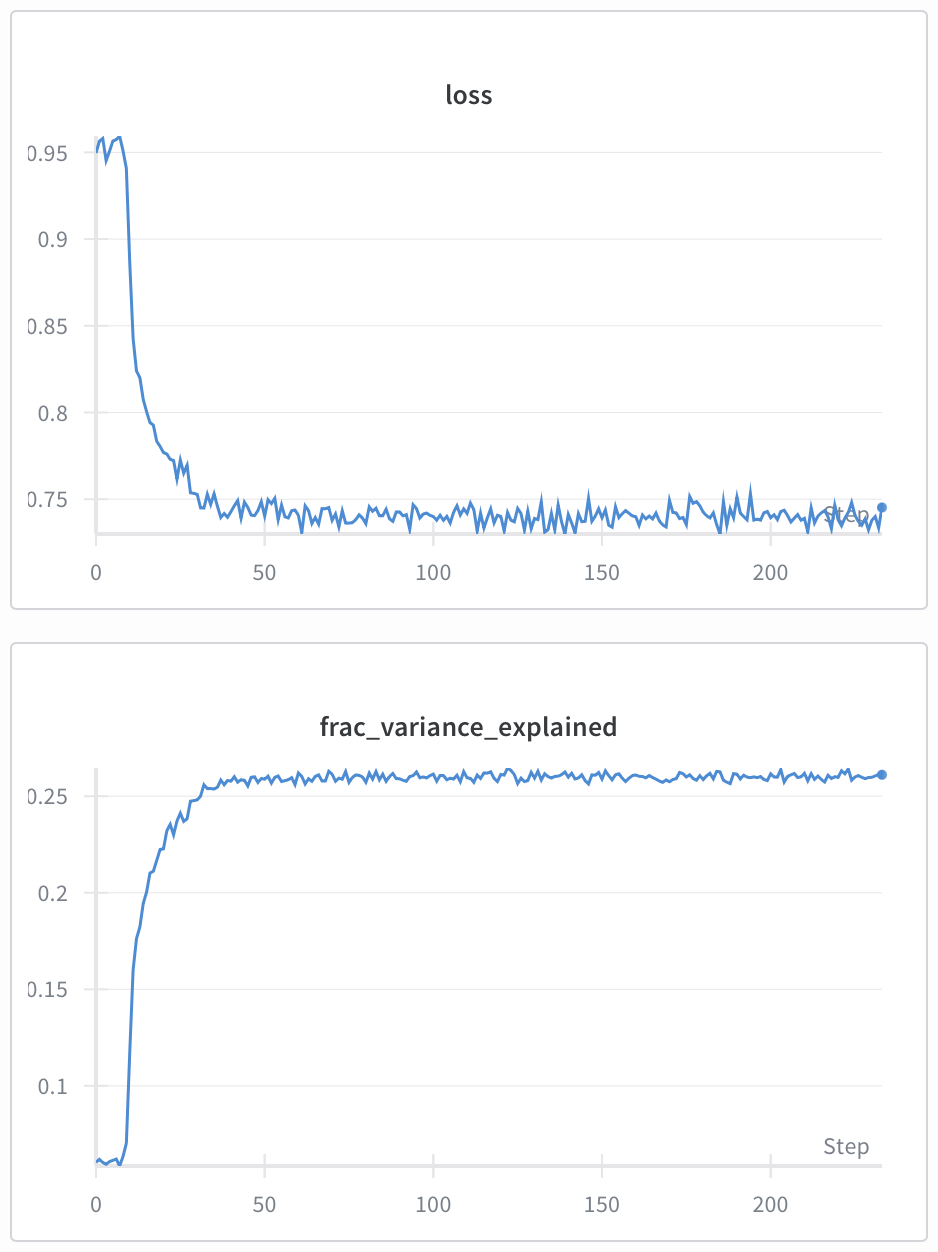}
    \caption{Typical Loss and FVE for Sampled-SAE on synthetic data}
    \label{fig:right}
  \end{subfigure}
  \caption{We train for 10,000 synthetic samples with training parameters same as the real data experiments.}
  \label{fig:synthetic_data_sae_train_plots}
\end{figure*}

Refer training plots Figure \ref{fig:synthetic_data_sae_train_plots}.


\subsection{Detailed K-Sparse Probing Results}
\label{K-sparse_probing_results}
Refer Table \ref{tab:probing-accuracy}.

\newcommand{\win}[1]{\textbf{#1}}
\newcommand{\winnum}[1]{{\bfseries #1}} 

\begin{table}[h!]
\centering
\caption{Comparison across datasets. Winners are selected by highest probe accuracy; ties are bolded. $\ell$ is shown inline with the architecture. FVU is lower-is-better.}
\label{tab:probing-accuracy}
\setlength{\tabcolsep}{5pt} 
\begin{tabular}{
  l 
  l 
  S[round-precision=3, table-format=1.3] 
  S[round-precision=4, table-format=1.4] 
}
\toprule
Dataset & Architecture (with $\ell$) & {FVU $\downarrow$} & {SAE acc $\uparrow$} \\
\midrule
\multirow{5}{*}{1}
  & SampledSAE (Entropy, $\ell$=3)        & 0.308 & 0.6822 \\
  & \wintext{SampledSAE (L2-norm, $\ell$=5)}   & 0.047 & \winnum{0.7434} \\
  & \wintext{SampledSAE (Leverage, $\ell$=5)}  & 0.047 & \winnum{0.7434} \\
  & SampledSAE (Uniform, $\ell$=4)        & 0.138 & 0.6876 \\
  & batch\_topk                       & 0.025 & 0.6758 \\
\midrule
\multirow{5}{*}{2}
  & SampledSAE (Entropy, $\ell$=3)        & 0.308 & 0.7276 \\
  & SampledSAE (L2-norm, $\ell$=5)        & 0.047 & 0.7612 \\
  & SampledSAE (Leverage, $\ell$=5)       & 0.047 & 0.7612 \\
  & SampledSAE (Uniform, $\ell$=4)        & 0.138 & 0.7214 \\
  & \wintext{batch\_topk}             & 0.025 & \winnum{0.7622} \\
\midrule
\multirow{5}{*}{5}
  & SampledSAE (Entropy, $\ell$=3)        & 0.308 & 0.7872 \\
  & \wintext{SampledSAE (L2-norm, $\ell$=5)}   & 0.047 & \winnum{0.8508} \\
  & \wintext{SampledSAE (Leverage, $\ell$=5)}  & 0.047 & \winnum{0.8508} \\
  & SampledSAE (Uniform, $\ell$=4)        & 0.138 & 0.7804 \\
  & batch\_topk                       & 0.025 & 0.8020 \\
\midrule
\multirow{5}{*}{10}
  & SampledSAE (Entropy, $\ell$=3)        & 0.308 & 0.8184 \\
  & SampledSAE (L2-norm, $\ell$=5)        & 0.047 & 0.8564 \\
  & SampledSAE (Leverage, $\ell$=5)       & 0.047 & 0.8564 \\
  & SampledSAE (Uniform, $\ell$=4)        & 0.138 & 0.8294 \\
  & \wintext{batch\_topk}             & 0.025 & \winnum{0.8764} \\
\bottomrule
\end{tabular}

\end{table}

\subsection{Unique Features}
\label{unique_features}
Refer Table \ref{tab:features unique to Entropy}.

\begin{table*}[h!]
\centering
\caption{Comparison between SampledSAE-Entropy latents and their best BatchTopK matches with explanations and similarity scores.}
\label{tab:features unique to Entropy}
\small
\begin{tabularx}{\textwidth}{c X c X c c c}
\toprule
\textbf{Entropy latent} & \textbf{Explanation} 
 & \textbf{Best-match latent} & \textbf{Explanation} 
 & \textbf{Overall} & \textbf{Decoder} & \textbf{Text} \\
\midrule
41150 & the word ``Let'' and the word ``Suppose'' in mathematical context 
      & 34460 & the term ``immunohistochemical'' and its variants in the context of biological analysis 
      & 0.092 & -0.063 & 0.247 \\
\addlinespace
27451 & pronouns and references related to female characters or subjects in various contexts 
      & 65343 & the term ``Visitor'' in programming contexts related to code structure and traversal 
      & 0.097 & -0.000 & 0.194 \\
\addlinespace
60055 & concepts related to low levels or minimum thresholds across various contexts 
      & 8195  & the name ``Robert'' in various contexts 
      & 0.112 & 0.067 & 0.158 \\
\addlinespace
52706 & specific acronyms or terms related to viruses and their associated RNAs 
      & 33561 & the terms related to the concepts of outer and inner structures or surfaces 
      & 0.128 & 0.041 & 0.214 \\
\addlinespace
56476 & the substring ``St'' in various contexts and technical terms 
      & 8195  & the name ``Robert'' in various contexts 
      & 0.136 & -0.022 & 0.294 \\
\bottomrule
\end{tabularx}

\end{table*}

\begin{table*}[h!]
\centering
\caption{Comparison between Squared-L latents and their best BatchTopK matches with explanations and similarity scores.}
\label{tab:features unique to Squared-L}
\small
\begin{tabularx}{\textwidth}{c X c X c c c}
\toprule
\textbf{Squared-$\ell$ latent} & \textbf{Explanation} 
 & \textbf{Best-match latent} & \textbf{Explanation} 
 & \textbf{Overall} & \textbf{Decoder} & \textbf{Text} \\
\midrule
18072 & terms involving variables and their coefficients in algebraic expressions 
      & 65343 & the term ``Visitor'' in programming contexts related to code structure and traversal 
      & 0.039 & 0.027 & 0.051 \\
\addlinespace
10307 & variables indicated by \texttt{<< >>} in mathematical expressions 
      & 53916 & phrases introducing examples or methods preceded by the words ``the following'' 
      & 0.064 & 0.007 & 0.120 \\
\addlinespace
13856 & references and citations related to MRI technology and findings in medical studies 
      & 15922 & legal terminology and biological concepts related to prostaglandins, sociodemographic factors, and interleukins 
      & 0.080 & 0.033 & 0.127 \\
\addlinespace
8118  & terms related to spectral analysis and data in scientific contexts 
      & 5340  & text related to the spinal cord and spinal injuries 
      & 0.094 & 0.099 & 0.090 \\
\addlinespace
27691 & legal case citations and references to court districts and decisions 
      & 15922 & legal terminology and biological concepts related to prostaglandins, sociodemographic factors, and interleukins 
      & 0.096 & 0.003 & 0.188 \\
\bottomrule
\end{tabularx}

\end{table*}

\begin{table*}[h!]
\centering
\caption{Comparison between L2 latents and their best BatchTopK matches with explanations and similarity scores.}
\label{tab:features unique to L2}
\small
\begin{tabularx}{\textwidth}{c X c X c c c}
\toprule
\textbf{L2-norm latent} & \textbf{Explanation} 
 & \textbf{Best-match latent} & \textbf{Explanation} 
 & \textbf{Overall} & \textbf{Decoder} & \textbf{Text} \\
\midrule
21059 & numbers and mathematical operations within calculations 
      & 53916 & phrases introducing examples or methods preceded by the words ``the following'' 
      & 0.024 & 0.048 & 0.000 \\
\addlinespace
29801 & attributes related to Android layout and styling elements 
      & 15922 & legal terminology and biological concepts related to prostaglandins, sociodemographic factors, and interleukins 
      & 0.082 & 0.035 & 0.129 \\
\addlinespace
50476 & the substring ``greater'' related to comparisons or quantities 
      & 33561 & the terms related to the concepts of outer and inner structures or surfaces 
      & 0.091 & -0.010 & 0.191 \\
\addlinespace
59710 & references to college education and experiences related to being a college graduate 
      & 5340  & text related to the spinal cord and spinal injuries 
      & 0.092 & 0.079 & 0.106 \\
\addlinespace
47382 & the phrase ``as'' followed by verbs or clauses that indicate explanations or transitions 
      & 53916 & phrases introducing examples or methods preceded by the words ``the following'' 
      & 0.098 & 0.016 & 0.181 \\
\bottomrule
\end{tabularx}

\end{table*}

\begin{table*}[h!]
\centering
\caption{Comparison between Uniform latents and their best BatchTopK matches with explanations and similarity scores.}
\label{tab:features unique to Uniform}
\small
\begin{tabularx}{\textwidth}{c X c X c c c}
\toprule
\textbf{Uniform latent} & \textbf{Explanation} 
 & \textbf{Best-match latent} & \textbf{Explanation} 
 & \textbf{Overall} & \textbf{Decoder} & \textbf{Text} \\
\midrule
1517 & expressions of gratitude and acknowledgment in responses or comments 
     & 8145 & terms indicating lack of limitations or conditions related to permissions and modifications 
     & 0.079 & 0.018 & 0.141 \\
\addlinespace
4574 & technical terms and URLs related to user guides and software documentation 
     & 33561 & the terms related to the concepts of outer and inner structures or surfaces 
     & 0.096 & 0.015 & 0.178 \\
\addlinespace
1688 & CSS properties and values related to layout and design elements 
     & 15922 & legal terminology and biological concepts related to prostaglandins, sociodemographic factors, and interleukins 
     & 0.099 & 0.037 & 0.160 \\
\addlinespace
5718 & chemical compounds and their structures or modifications in scientific contexts 
     & 8145 & terms indicating lack of limitations or conditions related to permissions and modifications 
     & 0.101 & 0.012 & 0.191 \\
\addlinespace
3943 & the phrase ``in order to'' indicating purpose or intention 
     & 8145 & terms indicating lack of limitations or conditions related to permissions and modifications 
     & 0.120 & 0.063 & 0.178 \\
\bottomrule
\end{tabularx}

\end{table*}

\subsection{Analysis of Top Features Across Architectures}
\label{app:feature-analysis}

Examining the Top-10 features by AutoInterp score (Tables~\ref{tab:top10-batchtopk}--\ref{tab:top10-uniform}) reveals how different scoring strategies implicitly select for different types of interpretable structure. BatchTopK's unrestricted selection yields the highest AutoInterp scores (0.887, Table~\ref{tab:top10-batchtopk}) but might include semantically overloaded features that combine unrelated concepts (e.g., feature 15922: "legal terminology, prostaglandins, sociodemographic factors"), suggesting it optimizes for reconstruction even at the cost of feature monosemanticity. L2-norm (Table~\ref{tab:top10-l2norm}) selects compositional building blocks—variables, HTML tags, API patterns—that likely combine to form complex representations, while Squared-$\ell$(Table~\ref{tab:top10-squared-ell}) similarly emphasizes relational operators ("more than", "greater") that structure information and tend to activate on various contexts. Entropy (Table~\ref{tab:top10-entropy}) captures discriminative patterns across granularities, from subword tokens ("St") to abstract concepts ("change/alteration"), suggesting it selects features that serve as semantic anchors. Uniform random selection (Table~\ref{tab:top10-uniform}) defaults to high-frequency syntactic elements (brackets, connectives), confirming that without guided selection, simpler surface patterns dominate. This distribution of feature types indicates that the "activation lottery" in BatchTopK may be a consequence of allowing all features—regardless of their computational role—to compete purely based on reconstruction error.

\begin{table}[h!]
\centering
\caption{Top-10 Features — BatchTopK (autointerp\_score = 0.887)}
\label{tab:top10-batchtopk}

\begin{adjustbox}{}
\begin{tabular}{@{}lllll@{}}
\toprule
Rank & Score & Latent & Explanation (short) \\ \midrule
1 & 1.0 & 15922 & legal terminology, prostaglandins, sociodemographic factors \\
2 & 1.0 & 5340  & spinal cord and spinal injuries \\
3 & 1.0 & 34460 & ``immunohistochemical'' in biological analysis \\
4 & 1.0 & 5865  & conditional compilation directives in code \\
5 & 1.0 & 53916 & phrases with ``the following'' introducing examples \\
6 & 1.0 & 33561 & outer/inner structures or surfaces \\
7 & 1.0 & 65343 & ``Visitor'' in programming contexts \\
8 & 1.0 & 8145  & absence of limitations/conditions (permissions) \\
9 & 1.0 & 8195  & the name ``Robert'' \\
10 & 1.0 & 8168 & legal evidence in judicial contexts \\ \bottomrule
\end{tabular}
\end{adjustbox}
\end{table}

\begin{table}[h!]
\centering
\caption{Top-10 Features — Entropy (l = 100.0, autointerp\_score = 0.821)}
\label{tab:top10-entropy}
\begin{adjustbox}{}
\begin{tabular}{@{}lllll@{}}
\toprule
Rank & Score & Latent & Explanation (short) & l \\ \midrule
1 & 1.0 & 27451 & pronouns/references related to female subjects & 100.0 \\
2 & 1.0 & 27129 & access modifiers and data types in programming & 100.0 \\
3 & 1.0 & 52706 & acronyms/terms related to viruses and RNAs & 100.0 \\
4 & 1.0 & 28279 & ``List'' in programming/data contexts & 100.0 \\
5 & 1.0 & 56476 & substring ``St'' in various contexts & 100.0 \\
6 & 1.0 & 4740  & the word ``first'' and initial occurrences & 100.0 \\
7 & 1.0 & 41150 & ``Let'' and ``Suppose'' in mathematics & 100.0 \\
8 & 1.0 & 60055 & concepts of low/minimum thresholds & 100.0 \\
9 & 1.0 & 1826  & concept of change/alteration & 100.0 \\
10 & 1.0 & 65040 & abbreviation ``U.S.'' in legal/patent contexts & 100.0 \\ \bottomrule
\end{tabular}
\end{adjustbox}
\end{table}

\begin{table}[h!]
\centering
\caption{Top-10 Features — Squared-$\ell$-norm (l = 100.0, autointerp\_score = 0.864)}
\label{tab:top10-squared-ell}
\begin{adjustbox}{}
\begin{tabular}{@{}lllll@{}}
\toprule
Rank & Score & Latent & Explanation (short) & l \\ \midrule
1 & 1.0 & 65439 & ``South'' and directional/geographical terms & 100.0 \\
2 & 1.0 & 18072 & variables and coefficients in algebra & 100.0 \\
3 & 1.0 & 27949 & HTML tags and document structure & 100.0 \\
4 & 1.0 & 55597 & requests, errors, and responses in programming & 100.0 \\
5 & 1.0 & 10307 & variables with $\ll \gg$ in mathematics & 100.0 \\
6 & 1.0 & 13856 & MRI technology references in medical studies & 100.0 \\
7 & 1.0 & 49697 & upload processes and object initialization in code & 100.0 \\
8 & 1.0 & 8118  & spectral analysis in scientific contexts & 100.0 \\
9 & 1.0 & 25824 & terms related to eye/vision (retina, eyelids) & 100.0 \\
10 & 1.0 & 27691 & legal case citations and court references & 100.0 \\ \bottomrule
\end{tabular}
\end{adjustbox}
\end{table}

\begin{table}[h!]
\centering
\caption{Top-10 Features — L2-norm (l = 100.0, autointerp\_score = 0.868)}
\label{tab:top10-l2norm}
\begin{adjustbox}{}
\begin{tabular}{@{}lllll@{}}
\toprule
Rank & Score & Latent & Explanation (short) & l \\ \midrule
1 & 1.0 & 48993 & circular shapes or domains in applications & 100.0 \\
2 & 1.0 & 35053 & thymus and histology (immune system) & 100.0 \\
3 & 1.0 & 23445 & ``document'' in programming/documentation & 100.0 \\
4 & 1.0 & 51553 & phrase ``more than'' (comparisons) & 100.0 \\
5 & 1.0 & 50476 & substring ``greater'' in comparisons & 100.0 \\
6 & 1.0 & 21059 & numbers and mathematical operations & 100.0 \\
7 & 1.0 & 29801 & Android layout and styling attributes & 100.0 \\
8 & 1.0 & 36472 & ``btn'' in HTML button elements & 100.0 \\
9 & 1.0 & 59710 & college education/graduate experiences & 100.0 \\
10 & 1.0 & 47382 & phrase ``as'' introducing explanations & 100.0 \\ \bottomrule
\end{tabular}
\end{adjustbox}
\end{table}

\begin{table}[h!]
\centering
\caption{Top-10 Features — Uniform (l = 100.0, autointerp\_score = 0.839)}
\label{tab:top10-uniform}

\begin{adjustbox}{}

\begin{tabular}{@{}lllll@{}}
\toprule
Rank & Score & Latent & Explanation (short) & l \\ \midrule
1 & 1.0 & 160   & substring ``Fab'' in names/terms & 100.0 \\
2 & 1.0 & 3943  & phrase ``in order to'' (purpose) & 100.0 \\
3 & 1.0 & 1517  & expressions of gratitude/acknowledgment & 100.0 \\
4 & 1.0 & 851   & ``\{'' indicating start of code blocks & 100.0 \\
5 & 1.0 & 2021  & programming errors and data structure declarations & 100.0 \\
6 & 1.0 & 1688  & CSS properties/values for layout/design & 100.0 \\
7 & 1.0 & 4474  & phrase ``due to'' (causality) & 100.0 \\
8 & 1.0 & 2607  & ``else'' in conditional statements & 100.0 \\
9 & 1.0 & 4574  & technical terms/URLs in documentation & 100.0 \\
10 & 1.0 & 5718  & chemical compounds and modifications & 100.0 \\ \bottomrule
\end{tabular}
\end{adjustbox}
\end{table}

\subsection{SAE Training Configurations}
\label{sae_training_config}
Refer Table \ref{tab:training-config}

\begin{table}[h!]
\centering
\caption{Training Configuration and Evaluation Setup}
\label{tab:training-config}
\begin{tabular}{@{}ll@{}}
\toprule
\textbf{Parameter} & \textbf{Value/Description} \\
\midrule
\multicolumn{2}{l}{\textit{Model and Data}} \\
Base Model & EleutherAI/pythia-160m-deduped \\
Training Layer & Residual stream after layer 6 \\
Activation Dimension & 768 \\
Context Length & 1024 tokens \\
Dataset & monology/pile-uncopyrighted (train split, streaming) \\
Training Tokens & 25M tokens (first 25M from The Pile) \\
\midrule
\multicolumn{2}{l}{\textit{SAE Architecture}} \\
Dictionary Size ($m$) & 65,536 \\
Effective L0 ($k$) & 60 \\
Candidate Pool Multiplier ($\ell$) & \{1, 3, 4, 5, 10, 20, 30, 40, 50, 60, 70, 100\} \\
& ($\ell=1$ gives $K$ candidates; $\ell=n/K$ recovers BatchTopK) \\
\midrule
\multicolumn{2}{l}{\textit{Training Hyperparameters}} \\
Training Steps & 50,000 \\
Batch Size & 4,096 tokens \\
Learning Rate & $3 \times 10^{-4}$ \\
Optimizer & Adam ($\beta_1$=0.9, $\beta_2$=0.999) \\
Gradient Clipping & 1.0 \\
Warmup Steps & 1,000 \\
Auxiliary Loss Weight ($\alpha$) & 1/32 \\
Threshold Start & Step 1,000 \\
Threshold Beta (EMA) & 0.999 \\
\midrule
\multicolumn{2}{l}{\textit{Initialization}} \\
Decoder & Unit-norm columns (maintained each step) \\
Encoder & Tied to decoder transpose \\
Bias ($b_{\text{dec}}$) & Geometric median of first batch \\
\midrule
\multicolumn{2}{l}{\textit{Sampling Configuration (SampledSAE)}} \\
Scoring Functions & \{Entropy, L2-norm, Squared-$\ell$, Uniform\} \\
Ridge Parameter ($\lambda$) & 0.01 (for Squared-$\ell$) \\
Feature Selection Frequency & Every batch \\
\bottomrule
\bottomrule
\end{tabular}
\end{table}

\end{document}